\documentclass[sn-apa]{sn-jnl}% APA Reference Style
\usepackage{multirow}
\usepackage{diagbox}
\usepackage{xcolor}
\usepackage{pifont} %for \ding

\usepackage{array}
\newcolumntype{L}[1]{>{\raggedright\let\newline\\\arraybackslash\hspace{0pt}}m{#1}}
\newcolumntype{C}[1]{>{\centering\let\newline\\\arraybackslash\hspace{0pt}}m{#1}}
\newcolumntype{R}[1]{>{\raggedleft\let\newline\\\arraybackslash\hspace{0pt}}m{#1}}
\newcolumntype{M}[1]{>{\centering\arraybackslash}m{#1}}
\newcolumntype{O}[1]{>{\raggedleft\arraybackslash}m{#1}}
\usepackage{makecell}
\usepackage{diagbox}

\jyear{2021}%

\theoremstyle{thmstyleone}%
\newtheorem{theorem}{Theorem}%  meant for continuous numbers
\theoremstyle{thmstyletwo}%

\theoremstyle{thmstylethree}%
\newtheorem{definition}{Definition}%

\newcommand{\cm}{\textcolor{blue}{\ding{51}}}%
\newcommand{\xm}{\textcolor{red}{\ding{55}}}%
\begin{document}

\title[UnbiasedNets: A Dataset Diversification Framework for Bias Alleviation]{UnbiasedNets: A Dataset Diversification Framework for Robustness Bias Alleviation in Neural Networks}


\author*[1]{\fnm{Mahum} \sur{Naseer}}\email{mahum.naseer@tuwien.ac.at}

\author[1]{\fnm{Bharath Srinivas} \sur{Prabakaran}}\email{bharath.prabakaran@tuwien.ac.at}

\author[2]{\fnm{Osman} \sur{Hasan}}\email{osman.hasan@seecs.nust.edu.pk}

\author[3]{\fnm{Muhammad} \sur{Shafique}}\email{muhammad.shafique@nyu.edu}

\affil*[1]{\orgname{Technische Universit\"at Wien (TU Wien)}, \orgaddress{\city{Vienna}, \postcode{1040}, \country{Austria}}}

\affil[2]{\orgdiv{School of Electrical Engineering \& Computer Science (SEECS)}, \orgname{National University of Sciences \& Technology (NUST)}, \orgaddress{\street{Sector H-12}, \city{Islamabad}, \postcode{44000}, \country{Pakistan}}}

\affil[3]{\orgdiv{Division of Engineering}, \orgname{New York University Abu Dhabi (NYUAD)}, \orgaddress{\city{Abu Dhabi}, \country{United Arab Emirates}}}


\abstract{
Performance of trained neural network (NN) models, in terms of testing accuracy, has improved remarkably over the past several years, especially with the advent of deep learning.
However, even the most accurate NNs can be biased toward a specific output classification due to the inherent bias in the available training datasets, which may propagate to the real-world implementations. 
This paper deals with the \textit{robustness bias}, i.e., 
% by a NN exhibiting a significantly large robustness to noise for a certain output class, as compared to the remaining output classes. 
the bias exhibited by the trained NN by having a significantly large robustness to noise for a certain output class, as compared to the remaining output classes. 
The bias is shown to result from imbalanced datasets, i.e., the datasets where all output classes are not \textit{equally represented}. \newline
Towards this, we propose the \textit{UnbiasedNets} framework, which leverages K-means clustering and the NN’s noise tolerance to diversify the given training dataset, even from relatively smaller datasets. 
This generates balanced datasets and reduces the bias within the datasets themselves. 
To the best of our knowledge, this is the first framework catering to the robustness bias problem in NNs. 
We use real-world datasets to demonstrate the efficacy of the \textit{UnbiasedNets} for data diversification, in case of both binary and multi-label classifiers. 
The results are compared to well-known tools aimed at generating balanced datasets, and illustrate how existing works have limited success while addressing the robustness bias. 
In contrast, \textit{UnbiasedNets} provides a notable improvement over existing works, while even reducing the robustness bias significantly in some cases, as observed by comparing the NNs trained on the diversified and original datasets.}

 %4-6 keywords only


\maketitle

%%================================%%
%%    Section 1: Introduction.    %%
%%================================%%

\section{Introduction} \label{sec:intro}

Machine learning (ML)-based systems are becoming increasingly ubiquitous in today’s world, with their applications ranging from small embedded devices (like health monitoring in smartwatches~\citep{health}) to large safety-critical systems (like autonomous driving~\citep{auto-drive}). 
Their success is often attributed to the Neural Networks (NNs) deployed in these systems, which have the ability to learn and perform decision-making with a high accuracy, without being explicitly programmed for their designated task. 
Typically, these NNs are trained on large datasets, with tens to hundreds of thousands of input samples, using various supervised training algorithms. 
Testing accuracy is often the most commonly (and possibly the only) used metric to analyze the performance of these NNs.

This spotlights two major limitations: (a) there is a notable reliance on large, labeled datasets, obtaining which is a significant challenge for the ML community, especially for new use-cases, and (b) the trained NN may experience problems like robustness bias, i.e., the robustness of NN to noise is not the same across all output classes, which accentuate in the presence of noisy real-world data.

Even when large datasets are available, they may contain a significantly large number of samples from one output/decision class. 
For instance, the MIT-BIH Arrhythmia dataset~\citep{arythmia-dataset} contains a considerably larger number of normal ECG signals as compared to the ECG signals indicating a specific arrhythmia. 
Likewise, the IMDB-WIKI dataset~\citep{imdb-dataset} comprises mostly of Caucasian faces. 
The NNs trained on such datasets are, therefore, less likely to detect arrhythmia or non-Caucasian faces, with high confidence - the problem aggravates under noisy input setting. 
However, the number of inputs from each output class is not the only parameter that leads to an \textit{imbalanced dataset}.

\subsection{Motivating Example}
\begin{figure}[ht]
    \centering
    \includegraphics[width=\linewidth]{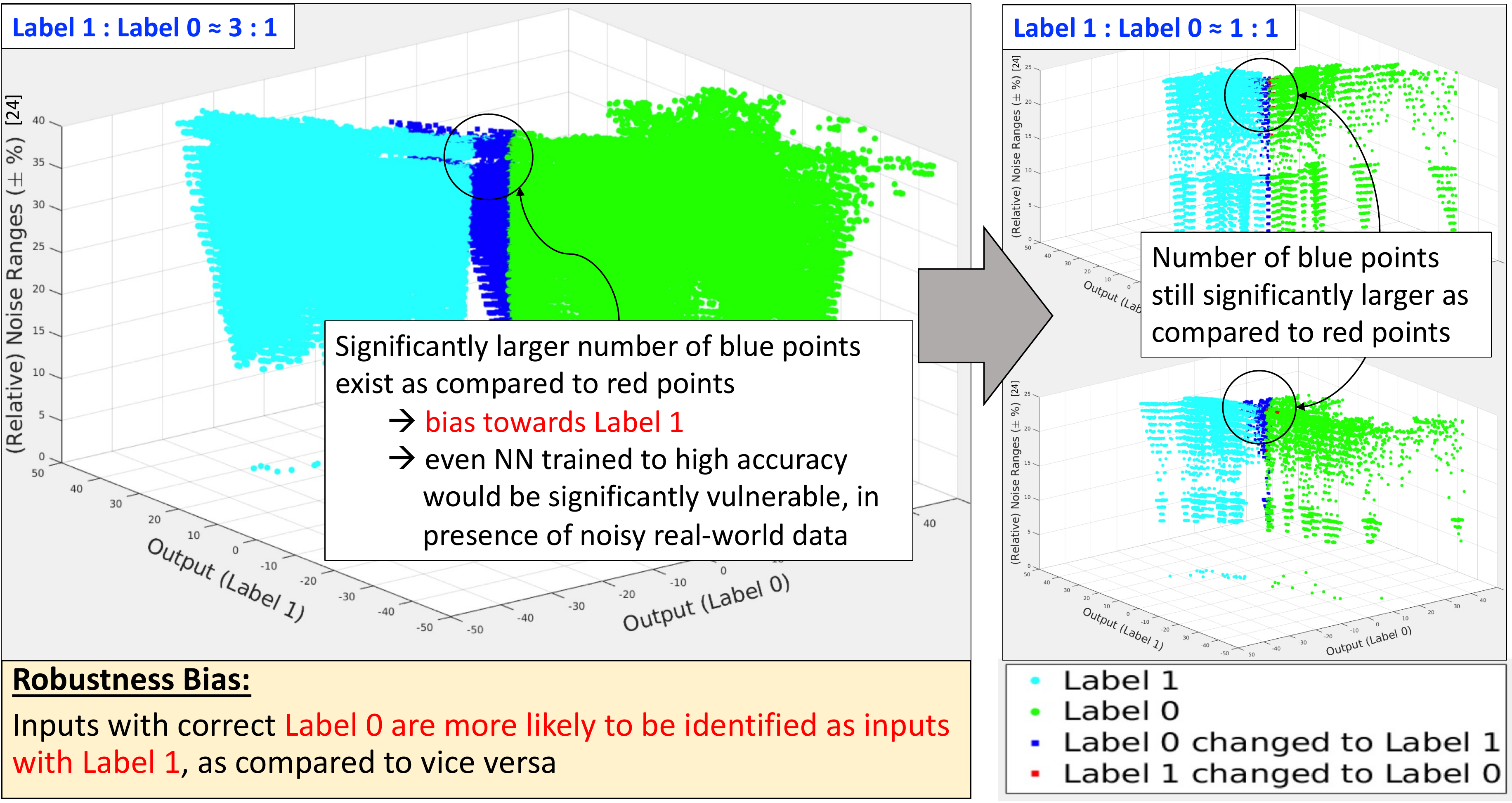}
    \caption{Networks trained on unequal (left) and equal (right) number of inputs from the classes: \textit{Label $0$} and \textit{Label $1$}. All networks used the same network architecture and training hyper-parameters, and all indicate a higher likelihood of \textit{Label $0$} being misclassified as compared to \textit{Label $1$}}
    \label{mot}
\end{figure}

Consider a NN trained on the Leukemia dataset~\citep{dataset} -- details of the dataset and NN are provided in Section \ref{sec:exp} along with further experiments.
The training dataset contains an unequal number of inputs from the two output classes. 
Fig.~\ref{mot} (left) shows the classification performance of this network under the application of varying noise. 
Not surprisingly, the trained NN is more likely to misclassify inputs from the output class with less number of training inputs.

The experiments were then repeated, deleting randomly selected inputs from the class with a larger number of inputs in the training dataset each time, hence ensuring an equal number of inputs from both classes in the dataset. 
The graphs in Fig.~\ref{mot} (right) give the classification performance of these networks under the application of varying noise. 
As shown in the graphs, simply having an equal number of inputs in both classes may still lead to a trained network significantly misclassifying inputs from one class. 

It must also be noted that the bias becomes apparent only in the presence of noise, since the trained NNs do not indicate misclassifications in the absence of noise. 
Hence, the robustness bias in a trained NN may go undetected before the deployment of the NN in a real-world application. This gravitates the need to address robustness bias and calls for the better description and acquisition of \textit{balanced datasets} that may enable training unbiased NNs. 
However, obtaining such datasets is not a straightforward task. 

The existing works dealing with bias alleviation either aim to improve the training algorithms to ensure unbiased training, or manipulate training data to obtain datasets that favor minimal NN bias. 
Yet, most of these works \citep{train2net-3,multi-modal,adversarialfilters} encounter the following limitations, making robustness bias alleviation a challenging task: 
\begin{enumerate}
    \item Most works \citep{repair,resound,constraints-training} focus on either the \textit{dataset bias}, i.e., the lack of generalization of the available dataset to real-world data, or \textit{representation bias}, i.e., flaws in the dataset acquired during its collection process. 
    However, they rarely focus on biases like \textit{robustness bias, which generally becomes evident only during NN deployment}, since noisy inputs are common in practical real--world systems.
    
    \item A \textit{limited notion of balanced dataset} is often used in literature \citep{comparison-toolboxused,comparison2021}, i.e., a balanced dataset is the one that contains an equal number of inputs from all output classes. However, as seen from our motivational example, such a dataset does not necessarily aid in the alleviation of robustness bias.
    
    \item They primarily focus on \textit{large datasets}~\citep{train2net-1,train2net-3,multi-modal,DataAug-training,adversarialfilters}, which provide a large pool of training samples to learn the input features from as well as to handpick a subset of inputs that favor an unbiased NN. However, such large datasets may not always be available.
    
    \item Some works focus on adding new input samples to the training dataset or at deeper network layers \citep{DataAug-training}. 
    However, the \textit{heuristics for adding new inputs} do not always favor a \textit{balanced dataset}.
    
    \item The addition and deletion of input samples \citep{comparison2021} may also lead to \textit{overfitting or reduction} of the training dataset, respectively.
    
    \item The works also often focus on \textit{visual datasets}, like colored MNIST or the IMDB dataset, where the existence of bias is perceptually easy to detect and comprehend~\citep{constraints-training,CIFARs}. 
    However, the \textit{robustness bias problem may stretch beyond visual datasets, albeit often being difficult to (perceptually) detect} in non-visual datasets.
\end{enumerate}

\subsection{Our Novel Contributions} 
To address the aforementioned limitations and challenges, this paper proposes the \textit{UnbiasedNets} framework\footnote{https://github.com/Mahum123/UnbiasedNets.git}, which facilitates the detection and reduction (ideally elimination) of bias in a trained NN by addressing the bias at the root level, i.e., by reducing the bias within the training data, rather than relying on training algorithms to unlearn biases. 
Our framework is generic and hence can be implemented along with any training algorithm, using any programming language (including MATLAB, Python, C++, etc.). The novel contributions of the work are as follows:

\begin{enumerate}
    \item This work deals with \textit{robustness bias}, which results from having an imbalanced dataset (which may in turn be a consequence of either dataset bias or representation bias or both), to alleviate bias from datasets where the bias may not always be apparent in the absence of noisy inputs.
    
    \item We redefine the notion of \textit{balanced dataset} to provide a more precise explanation of the extent to which the number of inputs from each output class is, or is not, essential for training unbiased NNs.
    
    \item Unlike the state-of-the-art approaches, \textit{UnbiasedNets} can work efficiently to diversify the dataset even in the \textit{absence of a large dataset} using K-means clustering and the noise tolerance of a NN previously trained on the dataset.
    
    \item Our novel framework can identify the practical bounds for generating synthetic input samples using clusters of input features obtained via K-means and the noise tolerance bounds of the trained network. 
    To the best of our knowledge, \textit{UnbiasedNets} is the only framework exploiting noise tolerance to obtain realistic bounds for synthetic inputs. 
    We also make use of feature correlation from real-world inputs to ensure that the \textit{synthesized inputs are realistic}.
    
    \item \textit{UnbiasedNets} combines synthetic input generation with redundancy minimization to diversify and generate potentially balanced and equally-represented datasets, \textit{with not necessarily an equal number of inputs from all output classes}.
    
    \item The framework is applicable in diverse application scenarios. 
    We demonstrate this using \textit{UnbiasedNets} on two real-world datasets, where the \textit{bias in the dataset is not always visually detectable}, and hence may not be straightforward to address.
\end{enumerate}

\unnumbered\subsection{Paper Organization}
The rest of the paper is organized as follows. Section \ref{sec:rw} gives an overview of the existing works for bias alleviation in NNs. Section \ref{sec:prelim} elaborates on the notions of balanced datasets, robustness, robustness bias, metric for bias estimation and noise tolerance, while also providing the relevant formalism. Section \ref{sec:framework} then explains our novel data diversification framework, \textit{UnbiasedNets}, to alleviate robustness bias from the training dataset. Sections \ref{sec:exp} and \ref{sec:analysis}
show the application of \textit{UnbiasedNets} on real--world datasets, providing details of experiments, results, and analysis. Section \ref{sec:diss} discusses the open future directions for the improvements in data diversification for alleviating robustness bias. Finally, Section \ref{sec:conc} concludes the paper.

%%================================%%
%%    Section 2: Related Works.   %%
%%================================%%

\numbered\section{Related Work} \label{sec:rw}

This section provides an overview of the current state-of-the-art on reducing bias in NNs. 
The summary of state-of-the-art, including approach categorization, their predominant focus on non-visual datasets, and their comparison to our novel \textit{UnbiasedNets} approach, is given in Table~\ref{tb:SoA}. 
The bias alleviation approaches can be broadly classified into two major categories: (1)~unbiased training algorithms (i.e., algorithm-centric (AC) approaches), and (2)~bias reduction via dataset manipulation (i.e., data-centric (DC) approaches). 
Towards the end of the section, we also provide an overview of the current and on-going works targeting the recently discovered problem of robustness bias.

\begin{sidewaystable}
% \sidewaystablefn%
\begin{center}
% \begin{minipage}{\textheight}
\caption{\protect\centering Comparison of the state-of-the-art bias alleviation approaches with our proposed \textit{UnbiasedNets} framework}
\centering
\begin{tabular*}{\textheight}{@{\extracolsep{\fill}}L{7cm}C{1.2cm}C{1.2cm}C{1.5cm}C{1.5cm}C{1.5cm}C{1.6cm}@{\extracolsep{\fill}}}

\toprule
Recent Work & Small Dataset & Non-Visual Dataset & Approach & Dataset Aug./Del. & Leverages $\Delta x_{max}$ & $X_{new}$ Validation \\\midrule
\cite{train2net-2} & \xm & \xm & AC & N/A & \xm & N/A \\
\cite{multi-modal} & \xm & \xm & AC & N/A & \xm & N/A \\
\cite{train2net-1} & \xm & \xm & AC & N/A & \xm & N/A \\
\cite{repair} & \xm & \xm & AC & Del. & \xm & N/A \\
\cite{train2net-3} & \xm & \xm & AC & N/A & \xm & N/A \\
\cite{sanh2020learning} & \xm & \cm & AC & N/A & \xm & N/A \\
\cite{savani2020intra} & \xm & \xm & AC & N/A & \xm & N/A \\
\cite{constraints-training} & \xm & \xm & AC & N/A & \xm & N/A \\
\cite{xu_2021_fairtrain} & \xm & \xm & AC & N/A & \xm & N/A \\
\cite{benz_2021_fairtrain} & \xm & \xm & AC & N/A & \xm & N/A \\%\hline
\cite{DataAug-training} & \xm & \xm & AC+DC & Aug. & \xm & \xm \\ %\hline
\cite{chawla2002smote} & \cm & \cm & DC & Aug. & \xm & \cm \\
\cite{he2008adasyn} & \cm & \cm & DC & Aug. & \xm & \cm \\
\cite{comparison-toolboxused} & \cm & \cm & DC & \multicolumn{1}{M{1.5cm}}{Aug./Del.} & \xm & \cm \\
\cite{adversarialfilters} & \xm & \cm & DC & Del. & N/A & N/A \\
\cite{resound} & \xm & \xm & DC & Del. & N/A & N/A \\
\textit{\textbf{UnbiasedNets}} & \cm & \cm & DC & \multicolumn{1}{M{1.5cm}}{Aug.+Del.} & \cm & \cm \\\bottomrule  
\end{tabular*}

\begin{tablenotes}
      \footnotesize
      \item Aug.: Input Augmentation ~~~~Del.: Input Deletion ~~~~N/A: technique not applicable for scenario ~~~~$\Delta x_{max}$: Noise Tolerance ~~~~ $X_{new}$: Synthetic Data
\end{tablenotes}
\label{tb:SoA}
% \end{minipage}
\end{center}
\end{sidewaystable}

\subsection{Algorithm-Centric (AC) Approaches}
Training unbiased NN via AC approaches often involves splitting the network model into two separate but connected networks~\citep{train2net-1,train2net-2,train2net-3}. 
The first network aims at either identifying key input features or amplifying the bias present in the dataset. 
The second network, in turn, uses these features or accentuated bias to unlearn the bias from the network. 
Learning features at deeper NN layers during training for data augmentation~\citep{DataAug-training} has also been shown to aid unbiased training. 
In addition, knowledge of known biases in the dataset and a NN trained using standard cross-entropy loss has also been leveraged to develop a more robust NN \citep{sanh2020learning}. 
Other AC bias reduction approaches include the incorporation of additional constraints during training to guide the NN in order to avoid learning unwanted correlations in data~\citep{constraints-training}.

For biases specific to multi-modal datasets (like colored MNIST~\citep{train2net-1}, where the dataset contains two kinds of information: the colors and the numerals), the use of a training algorithm based on functional entropy is shown to perform better~\citep{multi-modal}. 
A recent work~\citep{repair} also explores inputs in the dataset to identify the weights\footnote{Note that the weights for encoding inputs in \citep{repair} are not same as the parametric weights of NN layers.} that the inputs must be encoded with before training, to successfully reduce the bias. 
The determination of invariants in inputs has also been proposed \citep{invariant} to enable unbiased training of a NN.
In addition, recent work \citep{savani2020intra} also explores algorithms where instead of training an unbiased network from scratch, a trained NN and dataset (not used during training) are used to fine-tune the network to be devoid of biases specific to a certain application. 

However, as indicated earlier, these works are tailored for minimizing data and representation biases, generally for large datasets. 
The biases are often explored in visual datasets. 
In contrast, NNs deployed in the real--world often also deal with non-visual inputs, like patient’s medical data, where the existence of a bias (even the data and representation biases) may not always be easy to detect and hence may go unnoticed. 
Hence, bias alleviation poses a challenge in cases where the detection of bias is beyond visual perception. 
Moreover, the exploration of robustness bias is a fairly new research direction, and hence, the success of these AC approaches for minimizing robustness bias remains largely unexplored. 

\subsection{Data-Centric (DC) Approaches}
The orthogonal direction to minimize bias is by manipulating the training dataset via DC approaches, to potentially eliminate the bias at its core. 
Among the simplest and most popular DC bias alleviation approaches are random over-sampling (ROS), i.e., random replication of inputs from the class with less number of input samples, or random under-sampling (RUS), i.e., random deletion of inputs from the class with a significantly larger portion of available inputs \citep{class-imbalance-survey,comparison2021}.
The idea is to obtain a dataset with an equal number of inputs from each class. 
However, RUS is known to reduce the number of input samples available for NN to learn, while ROS may lead to overfitting the training data. 

The synthetic minority over-sampling (SMOTE)~\citep{chawla2002smote} and adaptive synthetic sampling (ADASYN)~\citep{he2008adasyn} techniques provide an improvement over ROS by synthesizing new points in the class with less number of samples using the available inputs as reference for the synthesis of new input samples \citep{comparison-toolboxused}. 
However, the general assumption in these works is that having an equal number of inputs for each of the classes ensures a balanced dataset, and in turn ensures an absence of bias~\citep{side-chan-paper,comparison2021}. 
As such, the approaches deploy data manipulation for the output class with a smaller number of inputs only. 
As observed in the motivating example in Section \ref{sec:intro}, this assumption provides a limited notion of balanced datasets. 
In addition, neither do these works have the means to ensure if the new inputs generated in fact belong to the minority class (i.e., output class with less number of inputs), nor the sophistication to analyze the number of inputs required to be added to the class to alleviate bias. 

Other works explore heuristics to identify the inputs that must be removed from the training dataset~\citep{resound,adversarialfilters} for obtaining an unbiased NN. 
However, for most real--world applications, large labeled datasets may not always be available, except to a few tech giants.
This leaves limited scope for tasks relying on limited dataset for bias alleviation.

In summary, the DC approaches again focus on alleviating representation and data bias, i.e., the biases pertaining to faulty data acquisition and lack of data generalizing well to all output classes. 
Alleviation of robustness bias remains an unexplored research direction in the existing works. 
The notion of a balanced dataset often used in these works is too naive. 
For the approaches relying on the deletion of inputs from the training dataset, the approaches are ideal only for large datasets to ensure sufficient inputs remain for NN training.
For the augmentation approaches (like ROS, SMOTE and ADASYN), i.e., the approaches where synthetic inputs are added to the training dataset (henceforth referred to as \textit{data augmentation}), the location for the new inputs is chosen to be in the close proximity around existing ``randomly'' selected inputs. 
The new inputs may or may not be realistic for the real-world input domain. 
The validation of these generated synthetic inputs relies solely on them being a part of NN training, and how well the trained NN works with the testing dataset. 

\unnumbered\subsubsection{Bias and the focus on Visual Datasets} 
As highlighted in Section \ref{sec:intro}, NNs are deployed in a diverse range of applications. 
These include networks performing classification and decision-making tasks for visual inputs \citep{masked_face,auto_drive}. 
Yet, a large portion of NN applications, for instance, banking \citep{bank}, environmental forecast \citep{solar}, finance \citep{finance} and spam filtering \citep{spam}, accept non-visual inputs. 
However, most literature pertaining to bias analysis \citep{train2net-1,train2net-2,train2net-3,multi-modal,constraints-training,DataAug-training,resound,repair} focus (often solely) on NNs working on visual datasets -- this comes to no surprise since a bias in these datasets is visually perceptible to human analysts, who are inclined  to perceive visual queues better than the non-visual ones (for instance, consider the case of visual capture, where visual senses are observed to dominate over auditory senses \citep{welch1999meaning}). 

The NNs using non-visual inputs often deploy similar network architectures as those using visual inputs. 
Intuitively, these NNs are likely to be as biased as their counterparts used in visual applications.
Yet, the difficulty in perceiving the bias in non-visual datasets makes their bias analysis a scarcely explored research area, as evident in the lack of existing works in the domain. 

Such dominant focus on visual datasets is not unique to the study of bias but is, in fact, also observed in fields like visual analytics, where non-visual aspects of the system are transformed into visual aspects. 
For example, the neuron activations are presented graphically (visually) in the research on network interpretability \citep{va_interpretable} and security \citep{va_sec}, which enables problem identification (detection). 
This in turn motivates deeper research/solutions. 

\numbered\subsection{Current and Ongoing Efforts}
The vulnerability of NNs to robustness bias has only been recently discovered \citep{rbias}. 
Hence, the efforts to resolve this particular category of bias are still limited. 
Nevertheless, a few AC approaches have been proposed within the last year to alleviate such bias. 
This includes a multi-objective training algorithm, which ensures that the standard error (which dictates the classification accuracy of the networks) and boundary error (since the inputs from class(es) closer to the decision boundary are expected to be more vulnerable under noise) \citep{xu_2021_fairtrain} are minimal, thereby minimizing the bias.
However, later work \citep{holistic2022} comes to a contrary conclusion, i.e., even the inputs with the same distance to the classification boundary may have different vulnerabilities to the noise. 
A re-weighting approach has also been proposed \citep{benz_2021_fairtrain}, which aims to update parameter values during training whenever the accuracy of a particular output class deviates too much from the average accuracy of the network. 

Recent work \citep{benz_2021_fairtrain} also notes that the bias in the NNs exists due to the dataset (and its features) itself, rather than depending on the NN model or its optimization factors. 
Yet, to the best of our knowledge, no DC effort has been proposed to alleviate bias from the dataset itself. 
It is interesting to note that adversarial training, a popular approach found successful in ensuring the robustness of NN against noise (concept explained later in Section \ref{subsec:rob}), is found to aggravate the bias \citep{adv_training_bad2021}.

%%================================%%
%%    Section 3: Preliminaries.   %%
%%================================%%

\section{Preliminaries} \label{sec:prelim}

This section describes the notions and provides the relevant formalism for balanced datasets, robustness, robustness bias, bias estimation and noise tolerance \citep{naseer2019fannet,rbias}, which form the basis of \textit{UnbiasedNets}. The terminology and notations introduced in the section will be used throughout the rest of the paper.

\subsection{Balanced Datasets} \label{subsec:bd}
Contrary to the popular notion, i.e., a balanced dataset \citep{comparison-toolboxused,comparison2021} consists of an equal number of inputs from all output classes, we define \textit{balanced dataset} to be the dataset where all output classes are \textit{equally-represented}. 
\begin{definition}[Balanced Dataset] \label{def:balanceddata}
\textit{Given a dataset $X$ with $\mathcal{L}$ output classes (i.e., $Y_1,Y_2,...,Y_\mathcal{L}$), the dataset is said to be balanced/the output classes are equally represented iff density $\rho$ of inputs from each class in the input hyperspace is (approximately) equal, i.e., $\rho({Y_1}) \approx \rho({Y_2}) \approx ... \approx \rho({Y_\mathcal{L}})$. Note that density $\rho$ of input here refers to the average number of input samples contained within the unit hypervolume of the valid input domain for an output class.}
\end{definition}
This implies that a network trained on such a balanced dataset would potentially be equally likely to identify inputs from all the classes, without a bias (explained in Section \ref{subsec:bias}).

\subsection{Robustness} \label{subsec:rob}
Robustness is the property of NN that signifies how the application of noise $\Delta x$ to the inputs does not change what the trained NN originally learned about the inputs.
\begin{definition}[Robustness]\label{def1}
\textit{Given a trained network $N : X \to Y$, $N$ is said to be robust against the noise $\Delta x$ if the application of an arbitrary noise $\eta \leq \Delta x$ to the input $x \in X$ does not change network's classification of $x$, i.e., $\forall \eta \leq \Delta x: N(x + \eta) = N(x)$.}
\end{definition}
It must be noted that $x$ corresponds to inputs that the network $N$ does not originally misclassify, i.e., $N(x)$ corresponds to the true output class for input $x$. 
For the purpose of this work, we assume the noise $\eta$ to be bounded within the L$^\infty$ space around input $x$, with the radius of $\Delta x$ -- this is one of the most popular noise used in NN analysis literature. 
Nevertheless, it is fairly straightforward to opt for any other type of (L$^p$-norm bounded) noise for the framework. 

\subsection{Robustness Bias} \label{subsec:bias}
Section \ref{sec:intro} highlighted the well-studied NN biases in literature, i.e., data and representation bias. This paper instead deals with \textit{robustness bias} (henceforth referred to as only \textit{bias}) proposed by \cite{rbias} and \cite{fair_detection}, which is a property of the dataset where a specific output class may or may not be robust under the application of noise. More specifically, it can be defined as follows:

\begin{definition}[Robustness Bias] \label{def:bias}
\textit{Given a dataset $X$ with $\mathcal{L}$ output classes (i.e., $Y_1,Y_2,...,Y_\mathcal{L}$), and $\mathcal{D}_{Y_1},\mathcal{D}_{Y_2},...,\mathcal{D}_{Y_\mathcal{L}}$ as the of input sub-domain representing each output class. $X$ is said to exhibit robustness bias iff the sub-domains $\mathcal{D}_{Y_1},\mathcal{D}_{Y_2},...,\mathcal{D}_{Y_\mathcal{L}}$ are not equidistant from the decision boundary.}
\end{definition}
Naturally, the sub-domains $\mathcal{D}_{Y_1},\mathcal{D}_{Y_2},...,\mathcal{D}_{Y_\mathcal{L}}$ may be disjoint or overlapping. 
However, as long as the sub-domains are equidistant from the decision boundary, the dataset is said to be free from a robustness bias. 
A NN trained on such a dataset is said to be unbiased, since intuitively, for a NN with a decision boundary equidistant from all input sub-domains, all output classes must be equally robust to noise. 

However, given the large number of input features (forming an input hyperspace) in practical datasets, it is not easy to visualize the bias in the dataset itself. 
Hence, we define the notion of biased NN, which aids in identifying the robustness bias in the dataset via analyzing the NN trained on the dataset:

\begin{definition}[Biased Network]\label{def2}
\textit{Given a trained network $N : X \to Y$, $N$ is said to be biased if the application of an arbitrary noise $\eta \leq \Delta x$ to any (correctly classified) input from class $X_i \subset X$ does not change network's output classification, $\forall \eta \leq \Delta x, x_i \in X_i: N(x_i + \eta) = N(x_i)$. However, application of the same noise to any input from another class $X_j \subset X$ makes the network misclassify the originally correctly classified input from the class $\forall \eta \leq \Delta x, x_j \in X_j: N(x_j + \eta) \neq N(x_j)$.}
\end{definition}
It must be noted that even though unbiasedness (i.e., the property of a trained NN to be unbiased) and classification accuracy may intuitively seem similar, they are not identical. 
Obtaining an accurate NN involves identifying the decision boundary that \textit{separates} the output classes in the dataset. 
In contrast, obtaining an unbiased NN involves identifying a decision boundary that is \textit{equidistant} from all the sub-domains encapsulating the different output classes. 
The resulting unbiased network, in turn, may or may not have the highest classification accuracy. 
However, all the output classes will likely be equally robust to noise in an unbiased network.

\subsection{Metric for Robustness Bias} \label{subsec:metric}
In practice, it is often impossible to obtain a completely unbiased NN. Hence, a metric is required to quantify and analyze the bias in the network. 
Let $R_i$ be the ratio of misclassified to correctly classified inputs from class $i$, which defines the average tendency of inputs from output class $i$ to be misclassified. We define the metric to estimate robustness bias ($\mathcal{B}_R$) as follows:
\begin{equation*}
    \mathcal{B}_R = max(abs(R_i - \frac{\sum_{j\in \mathcal{L}\setminus i}R_j}{\mid\mathcal{L}\mid - 1}))
\end{equation*} 
where $\mathcal{L}$ is the set of all output classes.
Having a $\mathcal{B}_R$ of zero indicates an equal $R_i$ across all output classes, and therefore an unbiased NN. 
Consequently, larger $\mathcal{B}_R$ implies higher bias. 
It must also be noted that the (absolute) difference in ratios $R_i$ and $R_j$ is generally different across the different pairs of output classes. 
In order not to reduce (nullify) the impact of the differences (and hence that of the bias in the network), the maximum difference, rather than the average, is used to estimate the bias in NN. 

Contrary to the formal notion of robustness bias, as provided in Def. \ref{def:bias}, $B_R$ uses the inputs to quantify bias rather than the decision boundary of the NN. 
This is a viable approach since the exact decision boundary of the NN is often hard to visualize for the multi-dimensional input space. 
The metric $B_R$, instead, makes use of the measurable/quantifiable entity, i.e., the input classification, to estimate the bias. 
As stated earlier, the ratio $R_i$ provides the tendency of the boundary to misclassify the inputs from class $i$. 
This is compared to the average tendency of misclassification of inputs from the other network classes $R_j$ -- this is analogous to comparing the distance of inputs to the decision boundary for different classes. 
Hence, if the ratio $R_i$ for all classes is equal (analogously all classes are equidistant from the decision boundary), $B_R$ computes to zero. 
The NN is then ought to be unbiased.

\subsection{Noise Tolerance}
Similar to robustness, noise tolerance also checks the classification performance of a NN for inputs under the application of noise. However, it is a stronger property than robustness (i.e., noise tolerance to a specific noise implies robustness to the noise as well) such that it provides the bounds within which the addition of noise does not change the classification of the inputs by a trained NN.

\begin{definition}[Noise Tolerance]\label{def3}
\textit{Given a trained network $N : X \to Y$, noise tolerance is defined as the \textit{maximum} noise ${\Delta x}_{max}$, which can be applied to a correctly classified input $x \in X$ such that $N$ does not misclassify the input. Hence, for any arbitrary noise $\eta \leq \Delta x_{max}$, the application of noise to an input $x \in X$ does not change network's classification of $x$, i.e., $\forall \eta \leq \Delta x_{max}: N(x + \eta) = N(x)$.}
\end{definition}
Alternatively, noise tolerance can be viewed as the largest $\delta$-ball ($l^\infty$ norm ball) around the inputs, such that $\delta = {\Delta x}_{max}$ and any input within this ball is correctly classified by the NN.  
Consequently, this knowledge can in turn be used to estimate the region around seed inputs where the realistic synthetic inputs may reside and still be correctly identified by a trained NN.

%X: input domain
%Y: output domain
%X_i: subset of domain X belonging to class i
%x: elements of domain X
%x_i: input from class i
%L: output classes
%x^a: feature from input x 
%||1+2||_1 = |1|+|2|
%||{a,b,c}}}_inf = max(a,b,c)

%%================================%%
%%    Section 4: UnbiasedNets.    %%
%%================================%%

\section{\textit{UnbiasedNets}: Framework for Bias Alleviation} \label{sec:framework}

\begin{figure}[ht]
	\centering
	\includegraphics[width=\linewidth]{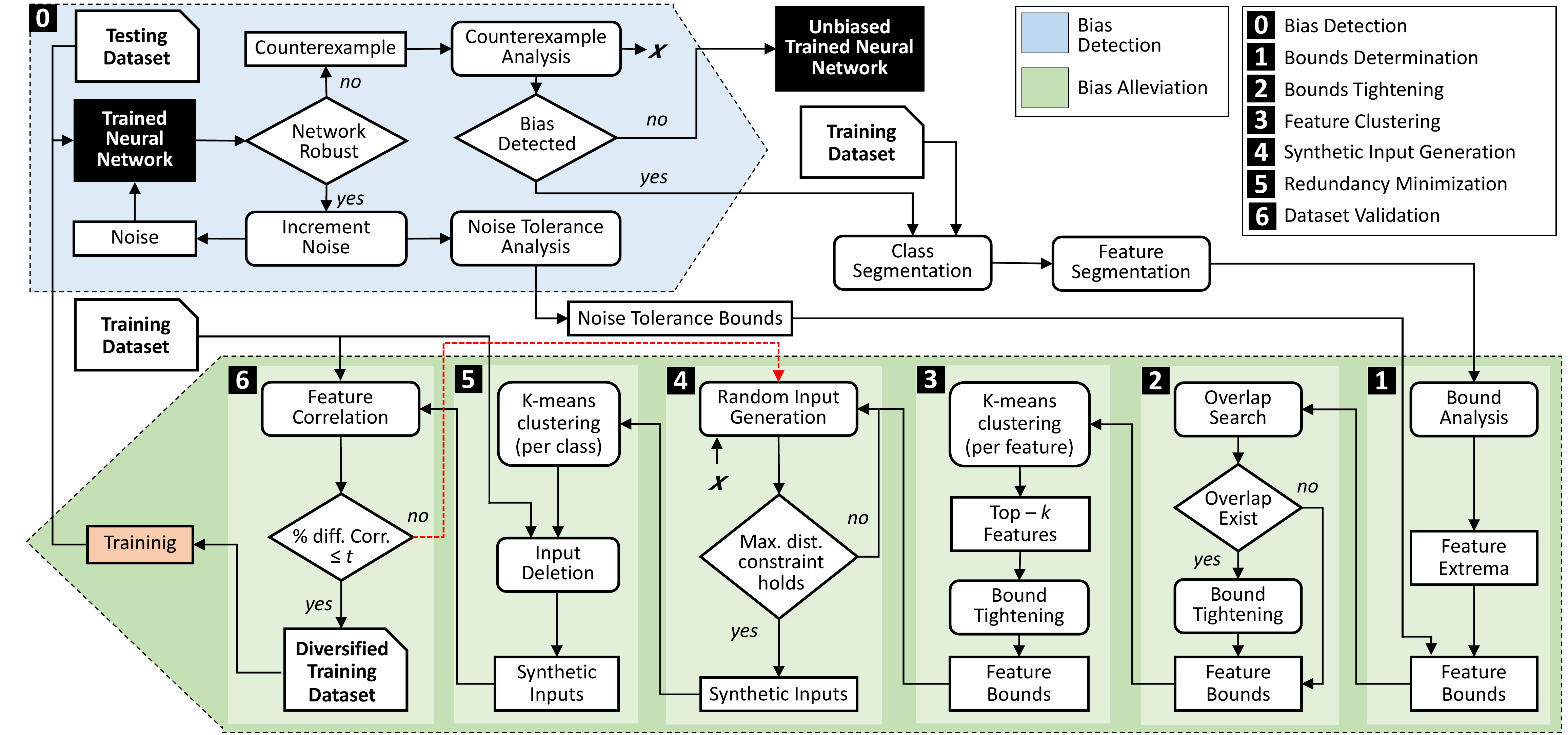}
	\caption{Overview of the \textit{UnbiasedNets} framework incorporating the proposed methodology starting with a trained NN undergoing bias detection, followed by bias alleviation, ultimately leading to a diversified dataset and potentially unbiased trained NN}
	\label{meth}
\end{figure}

We categorize \textit{UnbiasedNets} into two major tasks: \textit{bias detection} using a trained NN to identify the existence of robustness bias followed by \textit{bias alleviation} to diversify the training dataset to eliminate the bias at its core.
Fig.~\ref{meth} provides an overview of our proposed methodology.

\subsection{Bias Detection}

The first step here is the application of noise $\eta$, bounded by the small noise bounds $\Delta x$ to the inputs present in the testing dataset $x \in X$ (shown as Block $0$ in Fig.~\ref{meth}) to obtain the noisy inputs $x_n$. 
\begin{equation}
     x_n = x + \eta ~~~~s.t.~~~~ \eta\leq \Delta x
\end{equation}
The noisy inputs are then supplied to the trained NN, and their output classifications are compared to the classifications of inputs in the absence of noise. 
For the network to be robust (see Def.~\ref{def1}), the NN's classification must not change under the influence of noise. 
The noise is then iteratively increased, beyond the maximum noise at which the NN does not misclassify the inputs, i.e., beyond the NN's noise tolerance (see Def.~\ref{def3}). 
Such iterative increment of noise provides the noise tolerance bounds of the network.

The application of noise larger than the noise tolerance bounds of the NN entails that the NN  misclassifies some or all the noisy inputs. 
These misclassifying noise patterns (i.e., the counterexamples) act as inputs for the counterexample analysis. 
These noise patterns can be collected either using a formal framework (such as the ones based on model checking used by \cite{naseer2019fannet} and \cite{bhatti_probab}) or an empirical approach (like the Fast Gradient Sign Method (FGSM) attack \citep{FGSM}).

During counterexample analysis, the collected noise patterns, and in turn the misclassified inputs, are used to compute the $\mathcal{B}_R$ of the network to detect the presence and severity of robustness bias in the trained NN. 
A non-zero $\mathcal{B}_R$ implies a robustness bias in the network. 
Additionally, the number of misclassified inputs from each class is also used to determine the number of synthetic inputs required in the training dataset (elaborated in Section \ref{subsec:inp_gen}) to alleviate the bias.

\subsection{Bias Alleviation}

Using the noise tolerance available from the bias detection and the feature extremum of the inputs from the training dataset, we provide the step-by-step bias alleviation methodology.
The aim of the methodology is to identify the valid input domain for the generation of synthetic data and provide a diversified training dataset for the training of a potentially unbiased NN. 
The details of each step in the methodology are as follows:

\subsubsection{Bounds Determination} \label{subsec:bounds_determination}
\begin{figure}[h]
	\centering
	\includegraphics[width=\linewidth]{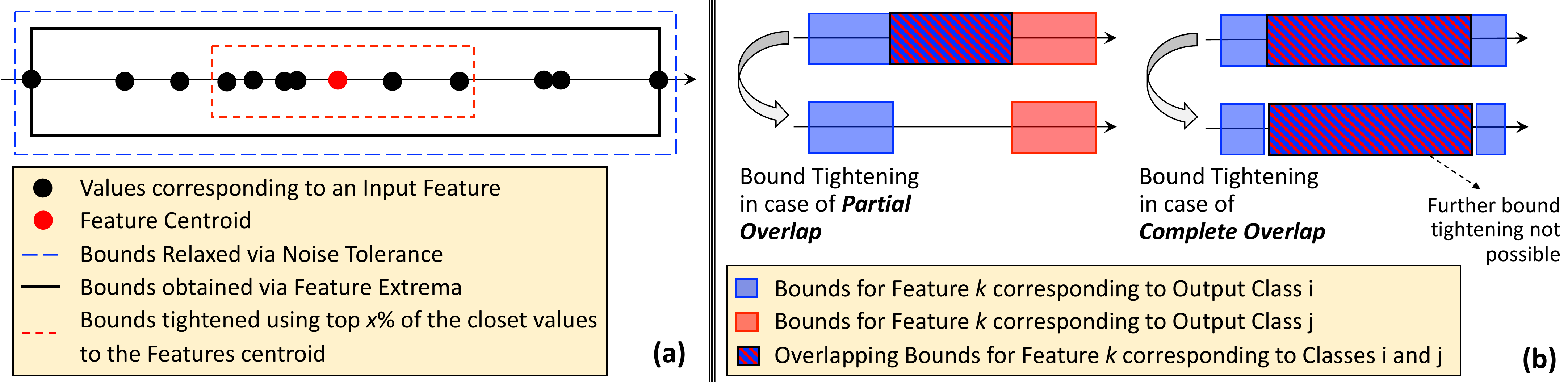}
	\caption{(a) Realistic bounds determination for individual feature bounds using available training inputs, K-means clustering and noise tolerance, (b) Bound tightening to eliminate/reduce bound overlap for synthetic input generation}
	\label{bound}
\end{figure}

For each input feature in every output class, the feature extremum, i.e., the maximum and minimum value of the feature as per the available training data, is first identified (as shown in Block $1$ of Fig.~\ref{meth}).
As discussed earlier, the inputs with noise, less than the allowed noise tolerance, are still likely to be correctly classified by a trained NN.
Hence, the feature bounds are relaxed using $\Delta x_{max}$, to provide a larger input space for the diversified inputs (also shown in Fig.~\ref{bound}(a)), as follows: 

\begin{theorem}[Bound Relaxation using Noise Tolerance]
\label{th1}
For input domain $X$, let $[\underline{x_i}, \overline{x_i}]$ represent the bounds of inputs belonging to $X_i$ (where $X_i \subset X$) and ${\Delta x}_{max}$ be the noise tolerance of the network. From Definition \ref{def2}, we know that the application of noise within the tolerance of the network does not change the output classification. Hence, more realistic input bounds $[\underline{x'_i}, \overline{x'_i}]$ can be obtained using the laws of interval arithmetic as:
\begin{equation*}
    % \begin{split}
    \underline{x'_i} = min((\underline{x_i}-\Delta x_{max}),(\underline{x_i}+\Delta x_{max}),(\overline{x_i}-\Delta x_{max}),(\overline{x_i}+\Delta x_{max})),
    % \end{split}
\end{equation*}
\begin{equation*}
    % \begin{split}
    \overline{x'_i} = max((\underline{x_i}-\Delta x_{max}),(\underline{x_i}+\Delta x_{max}),(\overline{x_i}-\Delta x_{max}),(\overline{x_i}+\Delta x_{max}))
    % \end{split}
\end{equation*}
\end{theorem}

It must be noted that due to the scalability of underlying bias detection framework (for instance \citep{naseer2019fannet}), where the application of large noise to NN inputs may lead to very large formal models, not suitable for analysis, noise tolerance may not always be available for bound relaxation. 
A similar challenge is encountered for NNs with a very low noise tolerance. 
Consider the example of a NN trained on an image dataset, where the addition of noise leading to a magnitude change of even $1.0$ in the pixel value of an image may still lead to misclassification \citep{small_noise}. 
This indicates a very low noise tolerance. 
Under these conditions, \textit{UnbiasedNets} assumes the noise tolerance to be zero, and proceeds with feature extremum as the feature bounds obtained during bound determination.

\subsubsection{Bound Tightening} 
Bounds obtained from the previous step identify the regions in the input space where real inputs from the training dataset exist, and hence provide an estimate for the generation of valid synthetic data.
However, it is possible for the feature bounds for different output classes to overlap, as shown in Fig.~\ref{bound}(b).
The overlap can be either \textit{partial} or \textit{complete}.
This provides a means for tightening the feature bounds (shown as Block $2$ in Fig.~\ref{meth}), hence leading to smaller, yet realistic, input space for the generation of synthetic data.
This in turn ensures that a lesser number of iterations are required for realistic synthetic input generation in the later steps of the framework.
The generation of tighter feature bounds in the case of partial feature can be seen as follows:

\begin{theorem}[Bound Tightening in case of Partial Overlap]
\label{th2}
Given the bounds of input feature $a$ for inputs belonging to class $i$ and $j$ to be $[\underline{x_i^a}, \overline{x_i^a}]$ and $[\underline{x_j^a}, \overline{x_j^a}]$, respectively, the bounds can be tightened to $[\underline{x_i^a}, \underline{x_j^a}]$ and $[\overline{x_i^a}, \overline{x_j^a}]$ provided that $\underline{x_i^a} < \underline{x_j^a}$ and $\overline{x_i^a} < \overline{x_j^a}$ (i.e., the bounds overlap partially). Then, any input belonging to the new bounds also belongs to the original feature bounds as well.
\begin{equation*}
    \begin{split}
        \forall i,j. (([\underline{x_i^a},\overline{x_i^a}] \in X_i^a \land &[\underline{x_j^a},\overline{x_j^a}] \in X_j^a ) \implies([\underline{x_i^a},\underline{x_j^a}] \in X_i^a \land [\overline{x_i^a},\overline{x_j^a}] \in X_j^a )) \\
        &s.t.~~  \underline{x_i^a} < \underline{x_j^a} < \overline{x_i^a} < \overline{x_j^a}
    \end{split}
\end{equation*}
\end{theorem}

However, the same cannot be generalized for complete overlap since the bounds of one label form a subset of the other. As such, tightening is possible for a single label only. 

\begin{theorem}[Bound Tightening in case of Complete Overlap]
\label{th3}
Given the bounds of input feature $a$ for inputs belonging to class $i$ and $j$ to be $[\underline{x_i^a}, \overline{x_i^a}]$ and $[\underline{x_j^a}, \overline{x_j^a}]$, respectively, the bounds for feature $a$ of class $i$, $X_i^a$, can be tightened to $[\underline{x_i^a}, \underline{x_j^a}]$ and $[\overline{x_j^a}, \overline{x_i^a}]$ provided that $\underline{x_i^a} < \underline{x_j^a}$ and $\overline{x_j^a} < \overline{x_i^a}$. Then, any input belonging to the new bounds for $X_i^a$ also belongs to the original feature bounds as well.
\begin{equation*}
    \begin{split}
        \forall i,j. (([\underline{x_i^a},\overline{x_i^a}] \in X_i^a \land &[\underline{x_j^a},\overline{x_j^a}] \in X_j^a ) \implies([\underline{x_i^a},\underline{x_j^a}] \in X_i^a \land [\overline{x_j^a},\overline{x_i^a}] \in X_i^a )) \\
        &s.t.~~  \underline{x_i^a} < \underline{x_j^a} < \overline{x_j^a} < \overline{x_i^a}
    \end{split}
\end{equation*}
\end{theorem}

~\\\noindent\textbf{\textit{Motivating Example.~}} Consider an arbitrary feature $a$ with valid input values in the range $[0,10]$. 
Let the inputs from class $i$ have the bounds $[2,8]$ and those from class $j$ have the bounds $[7,10]$, for the feature $a$. 
Without bound tightening, any input $7<x^a<8$ can belong to either class $i$ or $j$ (but not both). 
On the contrary, bound tightening reduces the bounds of the feature $a$ for classes $i$ and $j$ to $[0,7]$ and $[8,10]$, respectively. 
This reduces the valid input domain for feature $a$ such that it is impossible to pick a sample for feature $a$ that may belong to more than a single output class, hence simplifying the task of generating realistic synthetic input samples. 

\subsubsection{Feature Clustering} 
The previous steps in the framework make use of the entire training dataset to obtain realistic feature bounds.
But intuitively, real--world inputs often contain outliers that may be part of the training dataset, which do not occur frequently in practical case scenarios.
To subsume this characteristic into the synthetic inputs generated, further bound tightening is carried out (shown as Block $3$ in Fig.~\ref{meth}) on the \textit{top-k} input features, i.e., the $k$ features with the smallest distance from cluster centroid to the farthest input.

\subsubsection{Synthetic Input Generation} \label{subsec:inp_gen}
Using the feature bounds obtained from the previous step, the random input values are chosen within the available bounds (shown as Block $4$ in Fig.~\ref{meth}).
The number of inputs to be added to each output class $\chi_i$ is determined on the basis of the ratio of percentage of misclassified inputs from class $i$ (i.e., $\mu_i$) and the percentage of misclassified inputs from the class with minimum misclassifications (i.e., $min(\mu_L)$) using counterexamples recorded during the bias detection.
Hence, the class with higher $\mu_i$ gets the most synthetic inputs added to the dataset.

Algorithm \ref{algo1} outlines the entire synthetic data generation process, starting from the training dataset and noise tolerance bounds. 
Function \texttt{classSegment} (Line:~\ref{algo1ref1}) splits the dataset into non-overlapping subsets of inputs belonging to each class, \texttt{globalExt} (Line:~\ref{algo1ref2}) provides feature bounds using feature extremum, \texttt{nonOverlapping} (Line:~\ref{algo1ref3}) performs bound tightening on basis of Theorems \ref{th2} and \ref{th3}, \texttt{minDist} (Line:~\ref{algo1ref4}) identifies the \textit{top-k} features based on k-means clustering, \texttt{boundsFinal} (Line:~\ref{algo1ref5}) performs further bound tightening based on the top features, and \texttt{randInp} (Line:~\ref{algo1ref6}) finally generates the synthetic inputs for each output class.

\begin{algorithm}[h]{
			\footnotesize
			\caption{Synthetic Data Generation}
			\label{algo1}
			\begin{algorithmic}[1]
				\Statex \textbf{Input:} ~~Training Inputs ($X$), Number of Output Classes ($N$), Noise Tolerance ($\Delta x_{max}$), Number of top
				\Statex \tab ~~~~~~ Features to use for Bound Tightening $k$, Number of Inputs to add to each Class ($\chi$)
            	\Statex \textbf{Output:} Augmented Input Matrix ($X'$), Vector of Output Classes ($L'$)
\Statex

\Function{SynthGen}{$X,N,\Delta x_{max},\chi$}
\State $n=~$size($X$,$2$) \Comment{Number of Input Features}
\State ($X_1$,...,$X_N$)$~=~$classSegment($X$,$N$) \label{algo1ref1}

\For{$i=1$:$N$} 
\State $(f\_min_i,f\_max_i)~=~$ globalExt($\Delta x_{max}$,$X_i$)\label{algo1ref2} \Comment{Block $1$ in Fig.~\ref{meth}}
\EndFor

\For{$j=1$:$n$}
\State ($f'\_min^j$,$f'\_max^j$) $~=~$nonOverlappping($f\_min^j$ ,...,$f\_max^j$)\label{algo1ref3} \Comment{Block $2$ in Fig.~\ref{meth}}
\EndFor

\State ($T_1$,...,$T_k$)$~=~$minDist($X$)\label{algo1ref4}

\For{$m=1:k$}
\State ($f''\_min^{T_m}$,...,$f''\_max^{T_m}$)$~=$ boundsFinal($f'\_min^{T_m}$,$f'\_max^{T_m}$)\label{algo1ref5} \Comment{Block $3$ in Fig.~\ref{meth}}
\EndFor 

\For{$i=1$:$N$}
\State $X_{new}~=~$randInp($f''\_min_i$,...,$f''\_max_i$,$\chi_i$)\label{algo1ref6} \Comment{Block $4$ in Fig.~\ref{meth}}
\EndFor

\EndFunction
	\end{algorithmic}}
\end{algorithm}

\begin{figure}[ht]
    \centering
    \includegraphics[width=\linewidth]{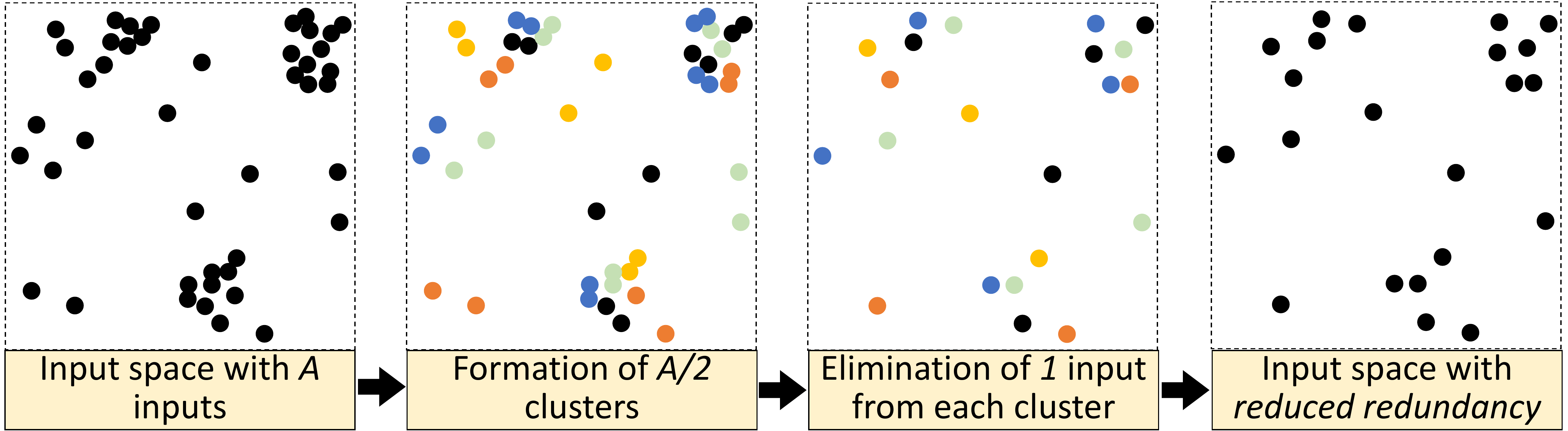}
	\caption{Redundancy minimization by $50\%$ in a two-dimensional input space}
    \label{red}
\end{figure}

It must be noted that the above input generation assumes an implicit hyperrectangular distribution of the input domain. This means, each input feature may take any input value (from within the defined input bounds), with equal likelihood. However, it is also possible for the input features to have non-rectangular distributions. Assuming these distributions to be known a priori, the random input generation could be modified to select input values, from within the input bounds, according to their probability of occurrence in their exact input distributions, i.e., with the more probable values having higher likelihood of selection  and vice versa.

\subsubsection{Redundancy Minimization} \label{subsec:redundant}
Oversampling may lead an NN to overfit to the training samples. 
Moreover, the existence of similar inputs, after the addition of synthetic inputs, does not add to the diversity of the dataset. 
Existing works also indicate that training the NNs on smaller datasets -- for instance, those obtained by eliminating input instances leveraging different distance matrices -- may reduce the timing overhead for training while providing comparable classification accuracy \citep{oversampling_overfitting_1,oversampling_overfitting_2,inst_select}. (Also see Appendix \ref{sec:appendix_ablation} for case studies indicating how redundancy minimization using K-means deletion reduces the bias of the actual NNs). 

Hence, $x\%$ closely resembling inputs from each class are removed to minimize the redundancy in the diversified training dataset (shown as Block $5$ in Fig.~\ref{meth}).
This is done by generating $\frac{1}{x}$ clusters for each output class and then retaining a single input from each cluster.
The result is a dataset with input samples covering diverse input space, without densely populating any specific region of the input space (as realized in Fig.~\ref{red}). 

\subsubsection{Dataset Validation} \label{subsec:dv}
Up until the previous step, \textit{UnbiasedNets} used real--world inputs to identify valid input space within which the inputs exist, used knowledge of the percentage of misclassified inputs from each output class to identify the number of synthetic inputs to generate, and minimized the redundancy in the generated input samples to obtain a diversified dataset.
However, features in the real--world data may be correlated, and the synthetic input features, despite lying in the valid input domain, may not follow the correlation of real--world data.
Hence, this step aims to validate the synthetic inputs by comparing their feature correlation with that of the original training data.
If the percentage difference between the correlation coefficients is within $t\%$, the new inputs are deemed suitable for training a potentially unbiased NN.
Otherwise, the process of synthetic data generation is repeated until the feature correlation of the synthetic inputs resembles that of the original training dataset (shown as Block $6$ in Fig.~\ref{meth}).

The choice of $t$ is made on the basis of the percentage difference between the correlation coefficients of training and testing datasets.
However, if this difference is too large, the features may simply be independent, or obtaining appropriate correlations may require some input pre-processing \citep{cross-corr}. 
The use of \textit{only} simple Pearson correlation coefficient, on such raw data, may not be an appropriate statistical measure to ensure the synthetic inputs to be realistic here. (Check Appendix \ref{sec:appendix} for more insights into this.)

%%================================%%
%%      Section 5: Experiments.   %%
%%================================%%

\section{Experiments} \label{sec:exp}

This section describes our experimental setup, and details of NNs and datasets used in our experiments.

\subsection{Experimental Setup}
All experiments were carried out on CentOS-7 system running on a $3.1$GHz $6$ core Intel i5-8600. 
Our \textit{UnbiasedNets} framework was implemented on MATLAB. The NN training was carried out using Keras.

However, the setup did not make use of any special libraries and, hence, can be easily re-implemented using any programming language(s).
Bias detection (and counterexample generation) was carried out using SMV models with applied noise in the range of $1 - 40\%$ of the actual input values, using a timeout of $5$ minutes for each input.

\subsection{Datasets and Neural Network Architecture} \label{subsec:dataset}
We experimented on the Leukemia dataset~\citep{dataset}, which is composed of the genetic attributes of Leukemia patients classified between Acute Lymphoblast Leukemia (ALL) and Acute Myeloid Leukemia (AML). 
The training dataset consists of $38$ input samples (with $27$ and $11$ inputs indicating ALL and AML, respectively), while the testing dataset contains $34$ inputs (with $20$ and $14$ ALL and AML inputs, respectively). 
We trained a single hidden layer ($20$ neurons), fully-connected ReLU-based NN, using the top-$5$ most essential genetic features from the dataset extracted using Minimum Redundancy and Maximum Relevance (mRMR) feature selection technique~\citep{matlab-leukemia}. A learning rate of $0.5$ for $40$ epochs followed by another $40$ epochs with a learning rate of $0.2$ were used during training.

We also experimented on the Iris dataset~\citep{iris,uci-repo}, which is a multi-label dataset, with characteristics of three iris plant categories as input features.
The dataset has an equal number of inputs from all output classes.
We split the dataset into training and testing datasets, with $120$ and $30$ inputs, respectively, while ensuring an equal number of inputs from all classes in each dataset.
A fully-connected ReLU-based two-hidden layer ($15$ neurons each) NN was trained with a learning rate of $0.001$ for $80$ epochs, using a training to validation split of $4:1$.

Since \textit{UnbiasedNets} is a data--centric bias alleviation framework, we compare the framework to well-acknowledged open-source state-of-the-art data--centric approaches: RUS, ROS, SMOTE~\citep{chawla2002smote} and ADASYN~\citep{he2008adasyn}. 
The Python toolbox \texttt{imbalanced-learn} implements all of the aforementioned techniques, except RUS, and was used for the generation of testing datasets. 
Since these approaches require the number of inputs to be different in each class, $50\%$ of the inputs from the Iris dataset were randomly selected to create a sub-dataset with an unequal number of inputs for the classes. 
RUS was implemented on MATLAB, removing inputs from class with more inputs to ensure both classes have the same number of inputs in the case of the Leukemia dataset and removing $25\%$ samples from each class in the case of the Iris dataset. 
To avoid overfitting during retraining of NNs using augmented datasets, the number of training epochs was reduced proportionally to the increase in the size of datasets. 

All NNs considered in the experiments were trained to the training and testing accuracies of over $90\%$. In addition, the experiments for each bias alleviation approach were repeated $10$ times to ensure conformity.

\section{Results and Analysis} \label{sec:analysis}
This section elaborates on the empirical results obtained from our experiments followed by comparison and analysis of \textit{UnbiasedNets} to the data-centric bias alleviation approaches.

\begin{table}[ht]
\begin{center}
\begin{minipage}{\linewidth}
\caption{Comparison of $\mathcal{B}_R$ values (average $\pm$ standard deviation) obtained for the NNs trained on original and diversified datasets, using open-source state-of-the-art approaches and \textit{UnbiasedNets} }

    \begin{tabular}{@{}lccc@{}}%{C{2.5cm}C{0.5cm}C{4.5cm}C{4.5cm}}
    \toprule
    \multirow{2}{*}{ \diagbox{Approach}{Datasets} }     & & \multicolumn{2}{c}{Robustness Bias ($\mathcal{B}_R$)} \\
    \cmidrule(r){3-4}
    & & Leukemia Dataset  & Iris Dataset                          \\
    \midrule
    Original   &   & \textbf{0.2228}  & \textbf{0.4732}   \\
    RUS  & & 0.1710 $\pm$ 0.07   & 0.5042 $\pm$ 0.11 \\
    ROS  & & 0.2213 $\pm$ 0.07 & 0.8059 $\pm$ 0.36 \\
    SMOTE & & 0.1452 $\pm$ 0.08 & 0.7709 $\pm$ 0.72\\
    ADASYN  &   & 0.2434 $\pm$ 0.06 & \textit{ADASYN not suited for dataset} \\
    \textit{UnbiasedNets}  &  & \underline{\textbf{0.1236 $\pm$ 0.05}}  & \underline{\textbf{0.4906 $\pm$ 0.15}}    \\ 
    \bottomrule
    \end{tabular}
    \label{tab:results}
\end{minipage}
\end{center}
\end{table}

% Leukemia + IRIS -> NuXmv
%MNIST -> Python

\subsection{Observations} 
As the number of output classes increases, ensuring an unbiased NN becomes a more challenging task. 
This was clearly observed in our experiments (Table~\ref{tab:results}), wherein the multi-label classifiers had a higher bias and at the same time, their bias reduction was substantially less effective in all bias alleviation approaches. 
(Note that the table represent the bias of the network trained on originaldataset is given in bold, and that of the network trained diversified dataset is given in bold italics.)

As discussed in Section~\ref{sec:prelim}, lower $\mathcal{B}_R$ indicates that the difference in the ratio of misclassified to correctly classified inputs is low, implying that the NN is less biased towards any output class. 
As summarized in Table~\ref{tab:results}, our \textit{UnbiasedNets} framework outperformed all the DC bias alleviation techniques while obtaining optimum $\mathcal{B}_R$ values for both binary and multi-label datasets. 
Moreover, in the case of the Iris dataset, using classical data--centric approaches to generate dataset with an equal number of inputs from each class seems to exacerbate the robustness bias. 
Although \textit{UnbiasedNets} may not always reduce the robustness bias, the data diversification ensures that the dataset remains balanced.
% Even RUS was found to make the NN more biased in case of the Iris dataset, as opposed to \textit{UnbiasedNets}, which reduces the same.

\begin{figure}[h]
	\centering
	\includegraphics[width=\linewidth]{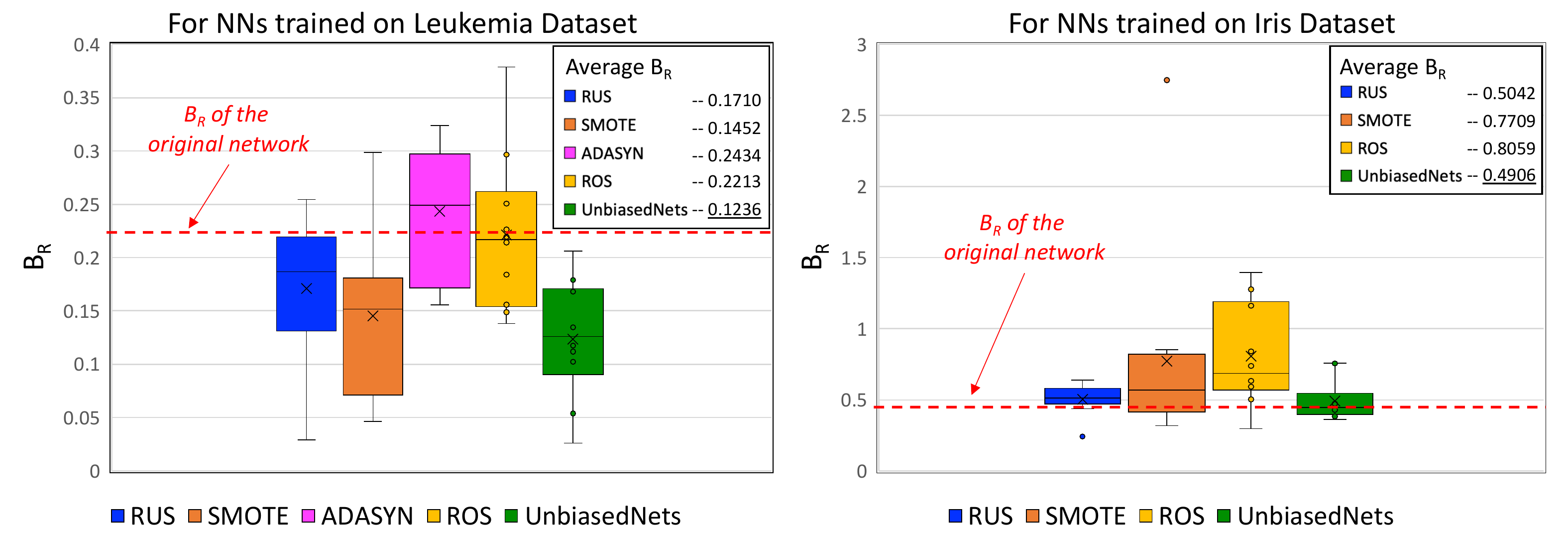}
	\caption{Variation in $\mathcal{B}_R$ results for NNs trained on RUS, SMOTE, ADASYN, ROS, and the diversified \textit{UnbiasedNets} datasets}
	\label{err}
\end{figure}

This success of biased can also be seen in Fig. \ref{err}, which shows the variation in $\mathcal{B}_R$ values over the repeated experiments. 
It is clearly evident that the individual experiments leading to a decrease in average robustness bias are far more compared to vice versa. 
Hence, we advocate executing several instances of experiments in order to obtain dataset instances that offer the best bias alleviation. 

Additionally, it can be seen from the box plots that NNs trained using the \textit{UnbiasedNets} datasets demonstrate considerably low interquartile ranges and the lowest average $\mathcal{B}_R$ values.
Even though RUS illustrates competitive $\mathcal{B}_R$ values, the use of RUS is not appropriate for small datasets, since the approach involves the deletion of real input samples and may hence diminish the learning capability of the NN.
The remaining approaches, i.e., ROS, SMOTE, and ADASYN, present a large variation in $\mathcal{B}_R$ results, deeming the approaches less effective for alleviation of robustness bias.

\subsection{Analysis} 
Our work focuses on robustness bias, which is exhibited by a trained NN in the presence of inputs having higher robustness to noise for certain output classes as compared to others. 
From our experiments%on NNs trained on original datasets
, we confirm the hypothesis that having an equal number of inputs (as in the case of Iris dataset) is in fact \textit{insufficient} to ensure an unbiased network.

In the case of the datasets where the number of inputs in each class is different, the known approaches like RUS, ROS, SMOTE, and ADASYN may reduce the bias. 
But for most datasets, they may be inadequate for robustness bias alleviation mainly for two reasons: 
(1) they rely on the naive definition of balanced datasets and only ensure the number of inputs for each class is equal, which overlooks the requirement of each class to be \textit{equally-represented} (concept explained in Section~\ref{subsec:bd}) in the input, and 
(2) during data augmentation, new inputs are only added in between the existing inputs, which neither diversifies the dataset sufficiently nor ensures that the new inputs are valid candidates for the augmented dataset. 
\textit{UnbiasedNets}, on other hand, uses counterexample analysis from the bias detection stage to obtain the required number of inputs in each class for a potentially \textit{equally-represented} dataset.
It also uses noise tolerance, which allows us to diversify the data beyond the bounds of the existing training dataset, which is subsequently validated by leveraging feature correlations, to alleviate bias in NN.

In the case of the Iris dataset, ROS and SMOTE were observed to significantly worsen robustness bias. 
This may be partially due to the deletion of inputs from the dataset to create an unequal number of inputs in the classes, which reduces the data available for NN training. 
However, RUS retained the $\mathcal{B}_R$ value close to the original dataset, even though the approach also employs input deletion. 
This suggests that the data augmentation by ROS and SMOTE may actually contribute to an exacerbation of bias rather than alleviation. 
In the case of \textit{UnbiasedNets}, even though the improvement in bias is often small, the results clearly suggest that diversifying the training dataset by adding realistic synthetic inputs and reducing redundancy in dataset is a potential direction to alleviate bias in NNs, unlike the other approaches. 

%%================================%%
%%     Section 6: Discussion.     %%
%%================================%%

\section{Discussion} \label{sec:diss}
\textit{UnbiasedNets} aims to diversify the dataset so as to (potentially) achieve a balanced dataset. 
While the diversification goal for obtaining a completely unbiased network may not always be achieved, \textit{UnbiasedNets} rarely aggravates the bias due to its precise perception of balanced datasets, unlike existing DC techniques. 
This section discusses the various aspects of NNs, which contribute to the challenge of data diversification and ultimately the persisting bias in trained networks. 

\subsection{Input Resemblance}
As seen from Table \ref{tab:results}, the greater the number of output classes, the higher the robustness bias in the NN. 
This implies that the higher the number of output classes, the more likely is the dataset imbalanced, and the more unlikely it is to obtain a trained NN that is equally robust for all output classes. 
A likely explanation for this could in fact be a close resemblance of inputs from the different classes, for datasets with a higher number of output classes. 
\begin{figure}[h]
	\centering
	\includegraphics[width=\linewidth]{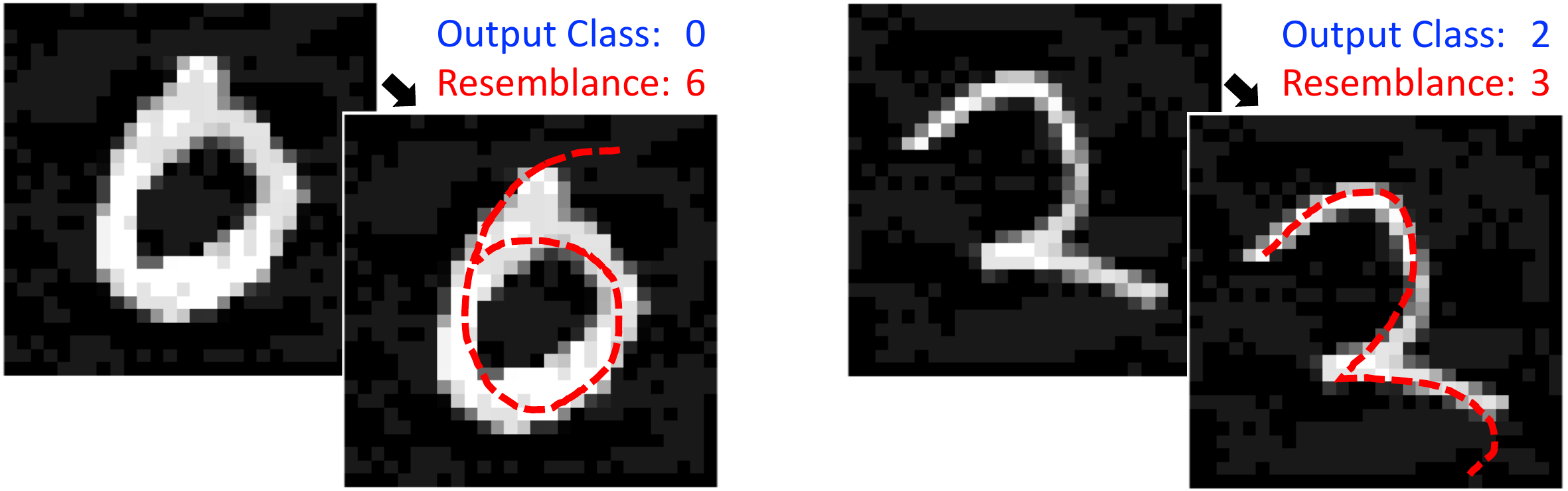}
	\caption{Inputs from one output class may resemble inputs from other classes, as observed in the MNIST dataset}
	\label{mnist-div}
\end{figure}

For instance, consider the case of hand-written digits (from the MNIST dataset), which comprises of $10$ output classes. 
As shown in Fig.~\ref{mnist-div}, it is possible for inputs from some classes to closely resemble inputs from other classes -- for example, digit $0$ may resemble a $6$, and digit $2$ may resemble a $3$. 
With inputs having likely resemblance to multiple classes, it is challenging to generate realistic synthetic inputs, and hence obtain successful data diversification for reducing the bias. 

A more careful study of the example provided above also reveals that the difference between the closely resembling inputs blur when their semantic distance is smaller \citep{semantics}, as shown in Fig.~\ref{mnist-div}. 
Yet, the syntactic rules for output classification stay intact even for these closely resembling inputs. 
For instance, a single loop forms the digit $0$, while an arc of a length comparable to half the circumference of the loop is required in addition to the loop to syntactically define the digit $6$. 
Hence, the addition of such syntactic rules for the generation of synthetic inputs (similar to the approach taken in neuro-symbolic learning \citep{neuro_symbolic}) may improve the data diversification. 

\subsection{Curse of Dimensionality}
Another challenge to data diversification is the large number of input neurons comprising the NN inputs -- a challenge often referred to as the ``curse of dimensionality" in the NN analysis literature \citep{survey2020}. 
This implies that as the number of input neurons for the NN increase, the computational requirements for its analysis increase exponentially. 

To understand this from the perspective of data diversification, let us consider the example of an image dataset. 
Data diversification determines input feature bounds directly from the raw input data to generate inputs such that the synthesized inputs $x$ belong to the valid input $\mathcal{D}$, i.e., $x\in \mathcal{D}$. 
However, various transformations, like affine, homographic and photometric transforms associated with image inputs may tremendously change the inputs, while still keeping the inputs realistic \citep{deepxplore}. 
Hence, for a practical image dataset, inputs belonging to even a single output class will have individual inputs that have undergone different transformations. 
As a result, the bounds of each input feature obtained from the inputs, for such a dataset, will be very large. 
This hinders the generation of synthetic data using these bounds, in turn making the data diversification halt at the data validation step since the search input space is too large for the randomly generated inputs to be realistic. (See Appendix \ref{sec:appendix} for details on the experimental analysis carried out to test the stated hypothesis on a real--world image dataset, MNIST.)

Towards this end, appropriate input pre-processing and the use of feature correlation knowledge to determine the bounds of the correlated input features (rather than raw input features) could potentially extend the applicability of \textit{UnbiasedNets} framework to a larger variety of datasets. 

%%================================%%
%%     Section 7: Conclusion.     %%
%%================================%%

\section{Conclusion} \label{sec:conc}

The overall performance of Neural Networks (NNs), particularly those relying on supervised training algorithms, is largely dependent on the training data available. 
However, the data used to train NNs may often be biased towards specific output class(es), which may propagate as robustness bias in the trained NN. 
But, unlike checking the testing accuracy of the trained NN, determining the bias in a NN is not a straightforward task. 
Existing works often rely on large datasets and aim at addressing biases by ensuring an equal number of inputs from each output class.
However, as shown by our detailed experiments, such approaches are not always successful. 
This paper proposes a novel bias alleviation framework \textit{UnbiasedNets}, which initially detects and quantifies the extent of bias in a trained NN and then uses a methodological approach to diversify the training datasets by leveraging the NN's noise tolerance and K-means clustering. 
To the best of our knowledge, this is the first framework specifically addressing the \textit{robustness bias} problem. 
We show the efficacy of \textit{UnbiasedNets}, using both binary and multi-label classifiers in our experiments, and also demonstrate how the existing bias alleviation may rather exacerbate the bias instead of alleviating it. 
We also discuss the challenges in robustness bias alleviation in certain datasets, and elaborate on the potential future research direction for addressing the robustness bias problem in trained NNs.

\backmatter

\bmhead{Acknowledgments}

This work was partially supported by Doctoral College Resilient Embedded Systems which is run jointly by TU Wien's Faculty of Informatics and FH-Technikum Wien, and partially by Moore4Medical project funded by the ECSEL Joint Undertaking under grant number H2020-ECSEL-2019-IA-876190. The authors also acknowledge TU Wien Bibliothek for financial support through its Open Access FundingProgramme.

\begin{appendices} 

\section{Ablation Studies} \label{sec:appendix_ablation}

\textit{UnbiasedNets} provide an overall framework for data diversification, leveraging K-means clustering and the noise tolerance of the trained NNs. 
As elaborated in Section \ref{sec:framework}, the methodology involves the perceptive addition of synthetic inputs to the existing dataset and consequently, minimizing the redundancy in the dataset using input deletion based on K-means clustering. 
This section provides ablation studies for both our novel synthetic input generation and deletion (i.e., redundancy minimization), to show the individual effectiveness of each component of the \textit{UnbiasedNets}. 
The studies make use of the NN trained on the Leukemia dataset (described in detail in Section \ref{subsec:dataset}). 
This is followed by a discussion to highlight the strengths and weaknesses of the components, motivating the sequential use of components, as adapted in our framework. 

\subsection{Synthetic Input Generation}
Here, the feature bounds are determined for individual input features using the bounds from the training dataset and the noise tolerance of the trained NN for synthetic input generation. 
These details of the process are elaborated in Sections \ref{subsec:bounds_determination}-\ref{subsec:inp_gen} and Algorithm \ref{algo1}. 
Synthetic inputs are generated for genetic attributes of AML leukemia (i.e., the output class with less number of samples in the training dataset). 
The updated dataset is validated (as elaborated in Section \ref{subsec:dv}) to ensure realistic input generation. 
Consequently, to avoid overfitting during training using this updated dataset, the number of training epochs is reduced to $56$, using the learning rates of $0.5$ and $0.2$ for $28$ epochs each, respectively. 

\subsection{Input Deletion}
In this study, the inputs are deleted from the output class with more input samples in the training dataset, as described in Section \ref{subsec:redundant}. 
The objective here is to leverage K-means clustering to reduce the redundancy in the training dataset, thereby potentially obtaining a balanced dataset for training. 
Again, the number of epochs used for training is updated to avoid overfitting, i.e., $69$ epochs are used each with the learning rates of $0.5$ and $0.2$, respectively, during training. 

\subsection{Results and Discussion}
The experiments based on studies provided earlier in the section were repeated $10$ times to ensure conformity. 
The resulting $\mathcal{B}_R$ from the experiments are summarized in Fig. \ref{fig:ablation}. 
It is clearly evident that both components of \textit{UnbiasedNets}, i.e., synthetic input generation and input deletion, aid in the reduction of bias in the trained network. 
\renewcommand{\thefigure}{7}
\begin{figure}[ht]
	\centering
	\includegraphics[width=\linewidth]{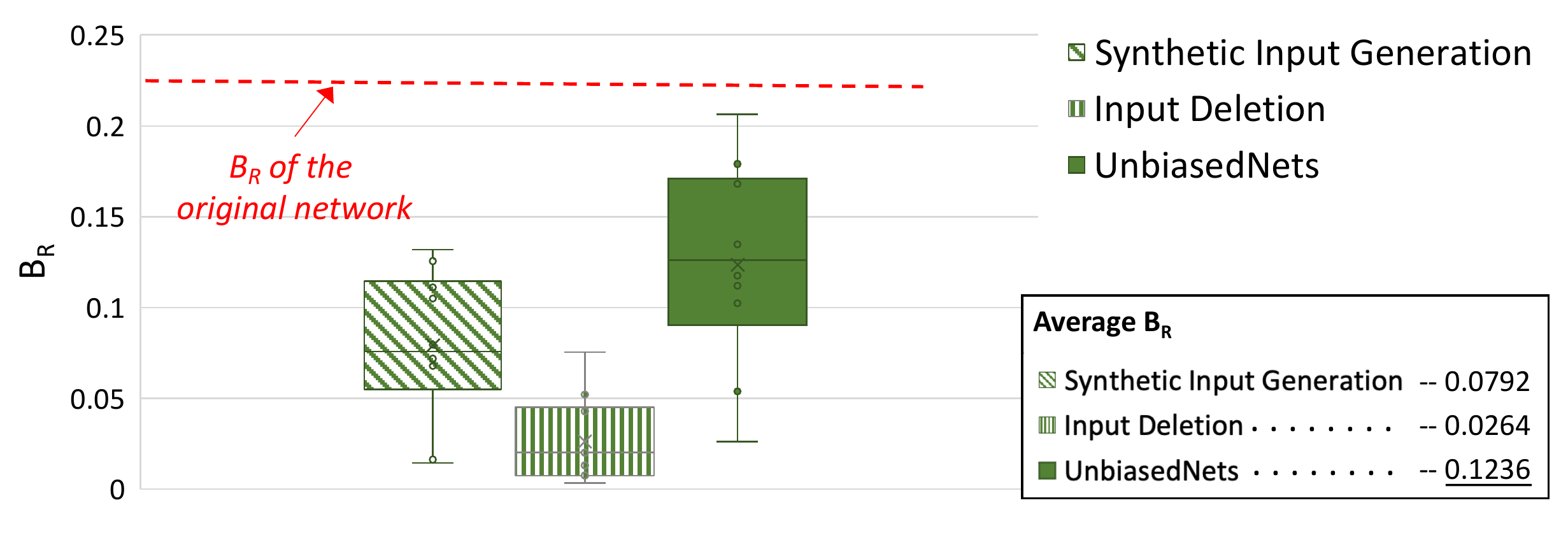}
	\caption{\textit{UnbiasedNets}, along with both its constituent components, i.e., synthetic input generation and input deletion, successfully diversifies the training dataset. This consequently reduces the bias in trained NN, as indicated by the lower $\mathcal{B}_R$ values.}
	\label{fig:ablation}
\end{figure}

Additionally, input deletion appears to provide the best overall reduction in the bias of the trained NN. 
However, it must be noted that the learning capability of NNs is data-driven, i.e., the more the training data available, the more likely is the trained NNs to perform well in real-world applications \citep{mayer2016large,xu2018towards}. 
Hence, to ensure optimal classification performance and generalization capabilities for trained NNs, the use of input deletion, standalone, is counter-intuitive. 

Bias reduction using synthetic input generation, on the other hand, appears only slightly better than that using \textit{UnbiasedNets}. 
However, since the selection of synthetic input samples is made using the existing data samples, it is possible for the new inputs to closely resemble the existing samples. 
Hence, not all synthetic inputs may add to the diversity of data. 

\textit{UnbiasedNets} leverages the strengths of both synthetic input generation (by providing a larger training dataset and potentially adding diversity to the data) and input deletion (by removing closely resembling input samples), to provide an overall data diversification framework proposed in Section \ref{sec:framework}. 
Hence, the framework not only diversifies the training dataset, but also reduces the bias in the trained NN (albeit not as significantly as its individual components), as observed in Fig. \ref{fig:ablation}. 

\section{Robustness Bias Alleviation for Image datasets} \label{sec:appendix}

To test the hypothesis (given in Section \ref{sec:diss}) that input feature bounds obtained directly from the raw input data are often too large for realistic synthetic input generation, we consider the problem of diversifying the MNIST (image) dataset. 
Details of the experiment, results, and analysis are as follows:

\subsection{Experimental Setup} 
We trained a LeNet-$5$ model on the MNIST dataset, which comprises of $60,000$ training and $10,000$ testing inputs, using Keras.
Training  and testing accuracies of $99.23\%$ and $98.78\%$ respectively, were achieved in $50$ epochs using a batch size of $1024$. 
The FGSM attack was implemented once for all inputs of the testing dataset using Adversarial Robustness Toolbox (ART)~\citep{art}. 
The adversarial noises were recorded for the counterexample analysis (as shown earlier in Fig. \ref{meth}) and ultimately bias detection.

\subsection{Results and Analysis} 
The number of inputs from each class in the MNIST testing dataset varies only slightly, as depicted by Fig.~\ref{mnist}(a). 
Yet, as highlighted in Section~\ref{sec:analysis}, the larger the number of output classes, the higher the chances of large robustness bias in the trained network, despite having an equal number of inputs from each class. 
This was observed in our neural network trained on the MNIST dataset (having $10$ output classes), which achieved a robustness bias $\mathcal{B}_R$ of $0.84$, as shown in Fig.~\ref{mnist}(b). 

\renewcommand{\thefigure}{8}
\begin{figure}[ht]
	\centering
	\includegraphics[width=\linewidth]{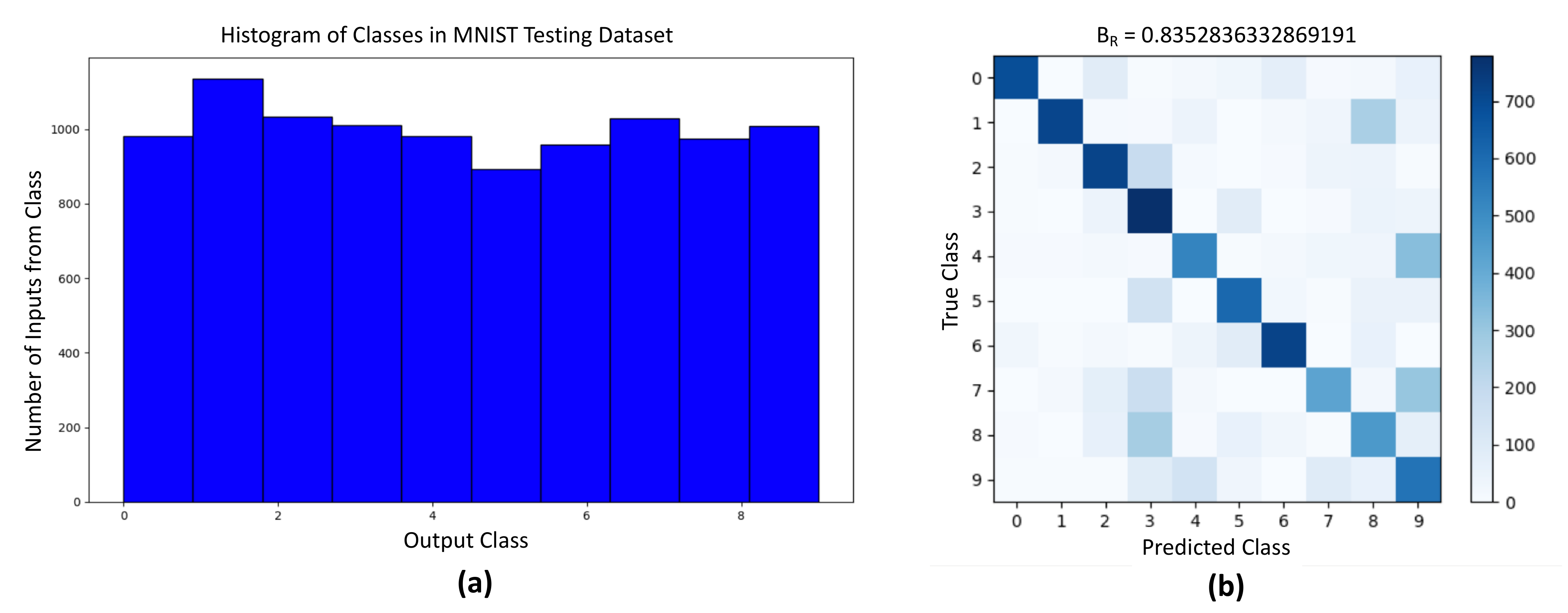}
	\caption{(a) Number of inputs from each class is approximately equal for MNIST's testing dataset, (b) the confusion matrix and $\mathcal{B}_R$ for LeNet-$5$ model trained on the original MNIST dataset, with FGSM attack used as a measure for bias detection}
	\label{mnist}
\end{figure}

Moreover, the percentage difference between the correlation coefficients of training and testing datasets in MNIST is in the factor of $10^4$. 
This is often evident in datasets like imaging datasets, which involve significant affine, homographic and photometric transformations leading to significantly different feature correlations in training and testing datasets. 
As hinted in Section \ref{subsec:dv}, this suggests that the input features are either independent or the input requires some pre-processing to obtain appropriate feature correlation. 

In the case of image inputs, the input features are already known to have spatial correlation \citep{corr}, albeit its determination is a non-trivial problem, particularly using only raw inputs \citep{cross-corr}. 
As expected, the resulting input feature bounds for the dataset are too large to enable the generation of realistic synthetic inputs to allow diversification. 
Hence, the use of appropriate input pre-processing, more sophisticated feature correlation measures \citep{cross-corr} and the inclusion of the correlated feature correlation knowledge during the bound determination process can potentially allow robustness bias alleviation for a wider range of machine learning datasets.



\end{appendices}

\bibliography{Ref}% common bib file

\end{document}